\pdfoutput=1

\documentclass[11pt]{article}

\usepackage{EMNLP2023}
\usepackage{times}
\usepackage{latexsym}

\usepackage[T1]{fontenc}
\usepackage[utf8]{inputenc}

\usepackage{amsmath}
\usepackage{color}
\usepackage{multirow,tabularx}
\usepackage{booktabs,longtable}
\usepackage{microtype}
\usepackage{graphicx}
\usepackage{bbding}
\usepackage{adjustbox}
\usepackage{stfloats}
\usepackage{enumitem}
\usepackage{verbatim}
\usepackage{amsfonts}
\usepackage{hyperref}
\usepackage{algorithm}
\usepackage{algorithmic}
\usepackage{subcaption}
\usepackage{caption}

\usepackage[all]{nowidow}

\title{Detecting Spoilers in Movie Reviews with External \\ Movie Knowledge and User Networks}

\newcommand{\ourmethod}[1]{}
\renewcommand{\ourmethod}[1]{\textsc{MVSD}}

\author{Heng Wang\textsuperscript{1} \ \ \ Wenqian Zhang \textsuperscript{2} \ \ \ Yuyang Bai\textsuperscript{1} \ \ \ Zhaoxuan Tan\textsuperscript{3} \\ \textbf{Shangbin Feng\textsuperscript{4}} \ \ \ \textbf{Qinghua Zheng\textsuperscript{1}} \ \ \ \textbf{Minnan Luo \textsuperscript{1}} \thanks{$^*$Corresponding author: Minnan Luo, School of Computer Science and Technology, Xi’an Jiaotong University, Xi’an 710049, China.}\\
  \textsuperscript{1}Xi'an Jiaotong University \ \ \ \textsuperscript{2}University of Manchester \\
  \textsuperscript{3}University of Notre Dame \ \ \ \textsuperscript{4}University of Washington  \\
\href{mailto:wh2213210554@stu.xjtu.edu.cn}{\texttt{wh2213210554@stu.xjtu.edu.cn}}}

\begin{document}
\maketitle
\begin{abstract}
Online movie review platforms are providing crowdsourced feedback for the film industry and the general public, while spoiler reviews greatly compromise user experience. Although preliminary research efforts were made to automatically identify spoilers, they merely focus on the review content itself, while robust spoiler detection requires putting the review into the context of facts and knowledge regarding movies, user behavior on film review platforms, and more. In light of these challenges, we first curate a large-scale network-based spoiler detection dataset \textbf{LCS} and a comprehensive and up-to-date movie knowledge base \textbf{UKM}. We then propose \textbf{\ourmethod{}}, a novel \textbf{M}ulti-\textbf{V}iew \textbf{S}poiler \textbf{D}etection framework that takes into account the external knowledge about movies and user activities on movie review platforms. Specifically, \textbf{\ourmethod{}} constructs three interconnecting heterogeneous information networks to model diverse data sources and their multi-view attributes, while we design and employ a novel heterogeneous graph neural network architecture for spoiler detection as node-level classification.  Extensive experiments demonstrate that \textbf{\ourmethod{}} advances the state-of-the-art on two spoiler detection datasets, while the introduction of external knowledge and user interactions help ground robust spoiler detection. Our data and code are available at \href{https://github.com/Arthur-Heng/Spoiler-Detection}{https://github.com/Arthur-Heng/Spoiler-Detection}.

\end{abstract}

\section{Introduction}

Movie review websites such as IMDB\footnote{\url{https://www.imdb.com}} and Rotten Tomato\footnote{\url{https://www.rottentomatoes.com}} have become popular avenues for movie commentary, discussion, and recommendation \citep{cao2019unifying}. Among user-generated movie reviews, some of them contain \emph{spoilers}, which reveal major plot twists and thus negatively affect people's enjoyment \citep{loewenstein1994psychology}. As a result, automatic spoiler detection has become an important task to safeguard users from unwanted exposure to potential spoilers.

\begin{figure}[t]
	\centering
	\includegraphics[width=1\columnwidth, height=4.42cm]{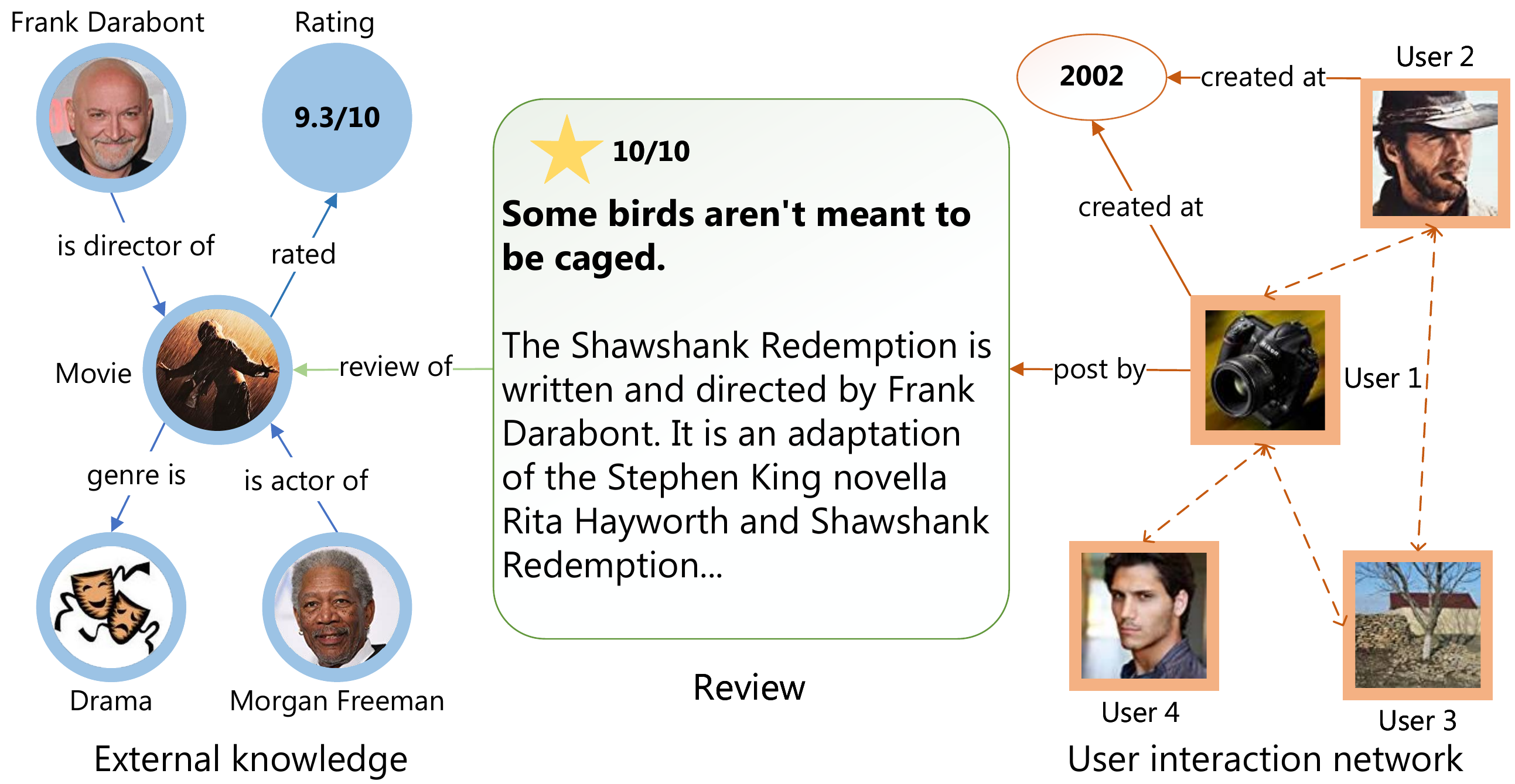}
	\caption{
        An example of a movie review and its context. 
        The review mentions Tim Robbins and Morgan Freeman, which are the names of the actors. Guided by external movie knowledge, the names can be recognized as the roles in the movie. Moreover, by incorporating user networks, it is discovered that User 1 likes to post spoilers on some specific genres of movies such as drama and comedy. Thus the review is more likely to be a spoiler.
	}
	\label{teaser}
\end{figure}

Existing spoiler detection models mostly focus on the textual content of the movie review. \citet{chang2018deep} propose the first automatic spoiler detection approach by jointly encoding the review text and the movie genre. \citet{wan-etal-2019-fine} extend the hierarchical attention network with item (i.e., the subject to the review) information and introduce user bias and item bias. \citet{chang-etal-2021-killing} propose a relation-aware attention mechanism to incorporate the dependency relations between context words in movie reviews. Combined with several open-source datasets \citep{10.5555/2655780.2655825,wan-etal-2019-fine}, these works have made important progress toward curbing the negative impact of movie spoilers.

However, robust spoiler detection requires more than just the textual content of movie reviews, and we argue that two additional information sources are among the most helpful for reliable and well-grounded spoiler detection. Firstly, \textbf{external knowledge} of films and movies (e.g. director, cast members, genre, plot summary, etc.) are essential in putting the review into the movie context. Without knowing what the movie is all about, it is hard, if not impossible, to accurately assess whether the reviews give away major plot points or surprises and thus contain spoilers. Secondly, \textbf{user activities} of online movie review platforms help incorporate the user- and movie-based spoiler biases. For example, certain users might be more inclined to share spoilers and different movie genres are disproportionally suffering from spoiler reviews while existing approaches simply assume the uniformity of spoiler distribution. As a result, robust spoiler detection should be guided by external film knowledge and user interactions on movie review platforms, putting the review content into context and promoting reliable predictions. We demonstrate how these two information sources can help spoiler detection in Figure \ref{teaser}.

In light of these challenges, this work greatly advances spoiler detection research through both resource curation and method innovation. We first propose a large-scale spoiler detection dataset \textbf{LCS} and an extensive movie knowledge base (KB) \textbf{UKM}. \textbf{LCS} is 114 times larger than existing datasets \citep{10.5555/2655780.2655825} and is the first to provide user interactions on movie review platforms, while \textbf{UKM} presents an up-to-date movie KB with entries of modern movies compared to existing resources \citep{misra2019imdb}. In addition to resource contributions, we propose \textbf{\ourmethod{}}, a graph-based spoiler detection framework that incorporates external knowledge and user interaction networks. Specifically, \textbf{\ourmethod{}} constructs heterogeneous information networks (HINs) to jointly model diverse information sources and their multi-view features while proposing a novel heterogeneous graph neural network (GNN) architecture for robust spoiler detection.

\begin{table}[t]
    \centering
    \caption{Statistics of LCS and existing dataset Kaggle.}
    \label{tab:compare_dataset}
    \begin{adjustbox}{max width=0.48\textwidth}
    \begin{tabular}{l c c c c}
         \toprule[1.5pt]
        \textbf{KB} & \textbf{\# Review} & \textbf{\# Cast} & \textbf{\# Metadata} &  \textbf{Year}\\
         
         \midrule[0.75pt]
        \textsc{Kaggle} & 573,913 & 0 & 5 & 2018 \\
        \textbf{LCS (Ours)} & 1,860,715 & 494,221 & 15 & 2022\\
         \bottomrule[1.5pt]
    \end{tabular}
    \end{adjustbox}
\end{table}

We compare \textbf{\ourmethod{}} against three types of baseline methods on two spoiler detection datasets.
Extensive experiments demonstrate that \textbf{\ourmethod{}} significantly outperforms all baseline models by at least 2.01 and 3.22 in F1-score on the Kaggle \citep{misra2019imdb} and LCS dataset (ours). Further analyses demonstrate that \textbf{\ourmethod{}} empowers external movie KBs and user networks on movie review platforms to produce accurate, reliable, and well-grounded spoiler predictions.


\section{Resource Curation}

We first curate a large-scale spoiler detection dataset \textbf{LCS} based on IMDB, providing rich information such as review text, movie metadata, user activities, and more. Motivated by the success of external knowledge in related tasks \citep{hu-etal-2021-compare, yao-etal-2021-knowledge, li-xiong-2022-kafsp}, we construct a comprehensive movie knowledge base \textbf{UKM} with important movie information and up-to-date entries.

\begin{table}[t]
    \centering
    \caption{Statistics of our proposed LCS dataset.}
    \begin{adjustbox}{max width=0.48\textwidth}
    \begin{tabular}{c|c|c}
        \toprule[1.5pt]  \textbf{Type}&\textbf{Number}&\textbf{Description}\\
        \midrule[0.75pt]
         review & 1,860,715 & The posting time is from 1998 to 2022. \\ 
         user & 259,705 & Users that posted these reviews. \\
         movie & 147,191 & The released year is from 1874 to 2022. \\ 
         cast & 494,221 & The cast related to the movies. \\ 
         spoiler & 457,500 & 24.59\% of the reviews are spoilers. \\
         \bottomrule[1.5pt]
    \end{tabular}
    \end{adjustbox}
    \label{tab:dataset}
\end{table}

\subsection{The LCS Dataset}
We first collect the user id of 259,705 users from a user list presented in the Kaggle dataset \citep{misra2019imdb}. We then retrieve the most recent 300 movie reviews by each user and collect the information of users, movies, and cast members based on the IMDB website. Since IMDB allows users to self-report whether its review contains spoilers, we adopt these labels provided by IMDB as annotations. We provide the comparison of our dataset to the Kaggle dataset in Table \ref{tab:compare_dataset}. As illustrated in Table \ref{tab:compare_dataset}, the LCS dataset has a much larger scale, more up-to-date information, and more comprehensive data. Details and statistics of the LCS datasets are presented in Table \ref{tab:dataset}.

\begin{table}[t]
    \centering
    \caption{Statistics of UKM and existing movie KBs.}
    \label{tab:benchmark}
    \begin{adjustbox}{max width=0.48\textwidth}
    \begin{tabular}{l c c c c}
         \toprule[1.5pt]
        \textbf{KB} & \textbf{\# Entity} & \textbf{\# Relation} & \textbf{\# Triple} & \textbf{Year}\\
         
         \midrule[0.75pt]
        \textsc{MovieLens} & 14,708 & 20 & 434,189 & 2019 \\
        \textsc{RippleNet} & 182,011 & 12 & 1,241,995 & 2018 \\
        \textbf{UKM (Ours)} & 641,585 & 15 & 1,936,710 & 2022\\
         \bottomrule[1.5pt]
    \end{tabular}
    \end{adjustbox}
\end{table}


\subsection{The UKM Knowledge Base}
Based on the LCS dataset, we then curate \textbf{UKM}, a comprehensive knowledge base of movie knowledge. 
We first assign each movie in the LCS dataset as an entity in the KB. We then collect all cast members and directors of these movies, de-duplicating them, representing each individual as an entity, and connecting movie entities with cast members based on their roles in the movie. After that, we further represent years, genres, and ratings as entities, connecting them to movie and cast member entities according to the information in the dataset.

We compare \textbf{UKM} against two existing movie knowledge bases (RippleNet \citep{wang2018ripplenet} and MoviesLen-1m \citep{cao2019unifying}) and present the results in Table \ref{tab:benchmark}, which demonstrates that \textbf{UKM} presents the largest and most up-to-date collection of movie and film knowledge to the best of our knowledge. \textbf{UKM} has great potential for numerous related tasks such as spoiler detection, movie recommender systems, and more.


\begin{figure*}[ht]
	\centering
	\includegraphics[scale=0.32]{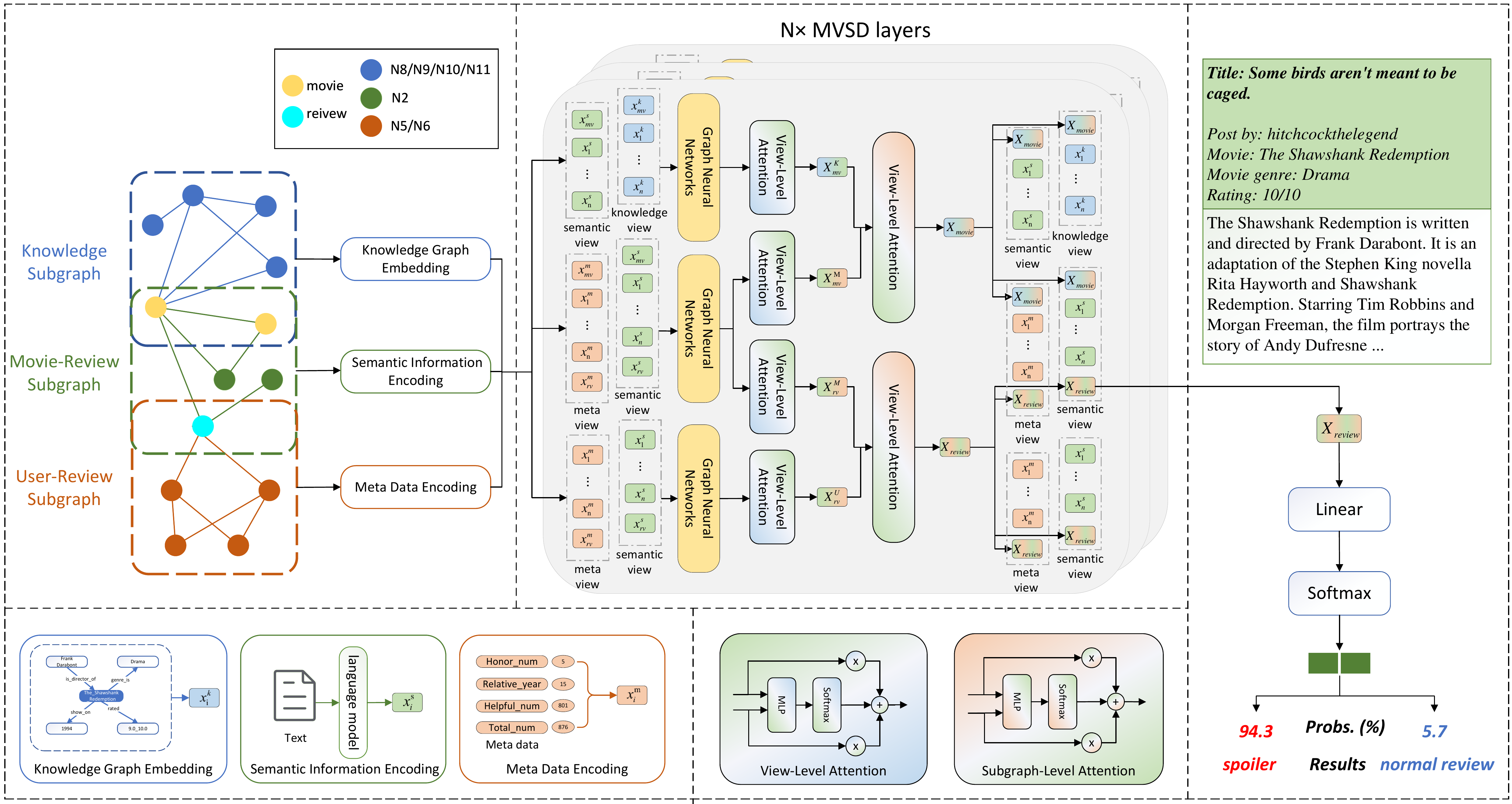}
	\caption{The architecture of \ourmethod{}, which incorporates external knowledge and social network interactions, leverages multi-view data and facilitates interaction between multi-view data.}
	\label{overview}
\end{figure*}

\section{Methodology}
We propose \ourmethod{}, a \textbf{M}ulti-\textbf{V}iew \textbf{S}poiler \textbf{D}etection framework. The overall architecture of the model is illustrated in Figure \ref{overview}. To leverage external movie knowledge and user activities that are essential in robust spoiler detection, \ourmethod{} constructs heterogeneous information networks to jointly represent diverse information sources. Specifically, we build three subgraphs: movie-review subgraph, user-review subgraph, and knowledge subgraph, each modeling one aspect of the spoiler detection process. \ourmethod{} first separately encodes the multi-view features of these subgraphs through heterogeneous GNNs, then fuses the learned representations of the three subgraphs through subgraph interaction. \ourmethod{} conducts spoiler detection with a node classification setting based on the learned representations of review nodes.
\subsection{Heterogeneous Graph Construction}
Graphs and graph neural networks have become increasingly involved in NLP tasks such as misinformation detection \citep{hu-etal-2021-compare} and question answering \citep{yu-etal-2022-kg}. In this paper, we construct heterogeneous graphs to jointly model textual content, metadata, and external knowledge in spoiler detection. 
Specifically, we first construct the three subgraphs modeling different information sources: movie-review subgraph $\mathcal{G}^K=\{\mathcal{V}^K,\mathcal{E}^K\}$, user-review subgraph $\mathcal{G}^M=\{\mathcal{V}^M,\mathcal{E}^M\}$, and knowledge subgraph $\mathcal{G}^U=\{\mathcal{V}^U,\mathcal{E}^U\}$.

\paragraph{Movie-Review Subgraph}
The movie-review subgraph models the bipartite relation between movies and user reviews. We first define the nodes denoted as $\mathcal{V}^M$, which include:

\noindent\underline{N1: \textit{movie}} The information about movies, especially the plot, is essential in spoiler detection. We use one node to represent each movie.

\noindent\underline{N2: \textit{rating}} Rating is an essential part of movie review. We use ten nodes to represent the numerical ratings ranging from 1 to 10.

\noindent\underline{N3: \textit{review}} We use one node to represent each movie review document.

We connect these nodes with three types of edges, denoted as $\mathcal{E}^M$:

\noindent\underline{R1: \textit{review-movie}} We connect a review node with a movie node if the review is about the movie.

\noindent\underline{R2: \textit{movie-rating}} We connect a movie node with a rating node according to the overall rating of the movie, rounded to the nearest integer.

\noindent\underline{R3: \textit{rating-review}} We connect a review node with a rating node based on its numeric score.

\paragraph{User-Review Subgraph}
The user-review subgraph is responsible for modeling the heterogeneity of user behavior on movie review platforms. The nodes in this subgraph, denoted as $\mathcal{V}^U$, include:

\noindent\underline{N4: \textit{review}} We use one node to represent each review document. Note that review nodes appear both in $\mathcal{V}^M$ (as N1) and $\mathcal{V}^U$ (as N4). Sharing nodes across subgraphs enables \ourmethod{} to model the interaction and exchange across different contexts.

\noindent\underline{N5: \textit{user}} We use one node to represent each user.

\noindent\underline{N6: \textit{year}} We use one node to represent each year, modeling the temporal distribution of spoilers.

We connect these nodes with three types of edges, denoted as $\mathcal{E}^U$:

\noindent\underline{R4: \textit{review-user}} We connect a review node with a user node if the user posted the review.

\noindent\underline{R5: \textit{review-year}} We connect a review node with a year node if the review was posted in that year.

\noindent\underline{R6: \textit{user-year}} We connect a user node with a year node if the user created the account in that year.

\paragraph{Knowledge Subgraph}
The knowledge subgraph is responsible for incorporating movie knowledge in external KBs. Nodes in this subgraph, denoted as $\mathcal{V}^K$, include:

\noindent\underline{N7: \textit{movie}} We use one node to represent each movie. 

\noindent\underline{N8: \textit{genre}} We use one node to represent each movie genre.

\noindent\underline{N9: \textit{cast}} We use one node to represent each distinct director and cast member.

\noindent\underline{N10: \textit{year}} We use one node to represent each year. 

\noindent\underline{N11: \textit{rating}} We use ten nodes to represent the numerical ratings ranging from 1 to 10. 

We connect these nodes with four types of edges:

\noindent\underline{R7: \textit{movie-genre}} We connect a movie node with a genre node according to the genre of the movie.

\noindent\underline{R8: \textit{movie-cast}} We connect a movie node with a cast node if the cast is involved in the movie.

\noindent\underline{R9: \textit{movie-year}} We connect a movie node with a year node if the movie was released in that year.

\noindent\underline{R10: \textit{movie-rating}} We connect a user node with a rating node according to the rating of the movie.\\

Note that the most vital nodes, movie nodes and review nodes, both appear in two subgraphs. These shared nodes then serve as bridges for information exchange across subgraphs, which is enabled by the MVSD model architecture in Section \ref{sec:MVSD}.

\subsection{Multi-View Feature Extraction}
The entities in the heterogeneous information graph have diverse data sources and multi-view attributes. In order to model the rich information of these entities, we propose a taxonomy of the views, dividing them into three categories.
\paragraph{Semantic View}
The semantic view reflects the semantics contained in the text.
We pass movie review documents, movie plot descriptions, user bio, and cast bio to pre-trained RoBERTa, averaging all tokens, and produce node embeddings $v^s$ as the semantic view.
\paragraph{Meta View}
The meta view is the numerical and categorical feature. We utilize metadata of user accounts, movie reviews, movies, and cast, and calculate the z-score as node embeddings $v^m$ to get the meta view. 
Details about metadata can be found in Appendix \ref{sec:metadata}.
\paragraph{Knowledge View}
The knowledge view captures the external knowledge of movies. Following previous works \citep{hu-etal-2021-compare, zhang-etal-2022-kcd}, we use TransE \citep{bordes2013translating} to train KG embeddings for the \textbf{UKM} knowledge base and use these embeddings as node features $v^k$ for the external knowledge view.

Based on these definitions, each subgraph has two feature views, thus nodes in each subgraph have two sets of feature vectors. Specifically, the knowledge subgraph $\mathcal{G}^K$ has the external knowledge view and the semantic view, the movie-review subgraph $\mathcal{G}^M$ and the user-review subgraph $\mathcal{G}^U$ has the meta view and the semantic view. We then employ one MLP layer for each feature view to encode the extracted features and obtain the initial node features $x_i^s$, $x_i^m$, $x_i^k$ for the semantic, meta, and knowledge view. 


\subsection{\ourmethod{} Layer}
\label{sec:MVSD}
After obtaining the three subgraphs and their initial node features under the textual, meta, and knowledge views, we employ \ourmethod{} layers to conduct representation learning and spoiler detection. Specifically, an \ourmethod{} layer first separately encodes the three subgraphs, then adopts hierarchical attention to enable feature interaction and the information exchange across various subgraphs.

\paragraph{\textbf{Subgraph Modeling}}
We first model each subgraph independently, fusing the two view features for each node. We then fuse node embeddings from different subgraphs to facilitate interaction between the three subgraphs. For simplicity, we adopt relational graph convolutional networks (R-GCN) \citep{schlichtkrull2018modeling} to encode each subgraph. For the $l$-th layer of R-GCN, the message passing is as follows:
\begin{equation*}
    \mathbf{x}_{i}^{(l+1)} = \Theta_{\textit{self}}  \cdot \mathbf{x}_{i}^{(l)} + \sum_{r \in \mathcal{R}} \sum_{j \in {\mathcal{N}_{r}}(i)} \frac{1}{|{\mathcal{N}_{r}}(i)|} \Theta_{r} \cdot \mathbf{x}_{j}^{(l)}
\end{equation*}
where $\Theta_{self}$ is the projection matrix for the node itself while $\Theta_{r}$ is the projection matrix for the neighbor of relation $r$.
By applying R-GCN, nodes in subgraph $\mathcal{G}^K$ get features from the knowledge and semantic view, denoting as $\mathbf{x}_k^K$ and $\mathbf{x}_s^K$, respectively. Nodes in subgraph $\mathcal{G}^M$ get features from the semantic and meta view, denoting as $\mathbf{x}_s^M, \mathbf{x}_m^M$, while nodes in subgraph $\mathcal{G}^U$ get the same views of feature, denoting as $\mathbf{x}_s^U, \mathbf{x}_m^U$.

\paragraph{\textbf{Aggregation and Interaction}}
Given the representation of nodes from different feature views, we adopt hierarchical attention layers to aggregate and mix the representations learned from different subgraphs. Our hierarchical attention contains two parts: view-level attention and subgraph-level attention. Considering movie node and review node are shared nodes of subgraphs and are of the most significance, we utilize these two kinds of nodes to implement our hierarchical attention.

We first conduct view-level attention to aggregate the multi-view information for each type of node. For each node in a specific subgraph, it has embeddings learned from two types of feature views. We first adopt our proposed view-level attention to fuse the information learned from different views for each node. We learn a weight for each view of features in a specific subgraph. Specifically, the learned weight for each view in a specific subgraph $\mathcal{G}$, $(\alpha_{v_1}^{\mathcal{G}}, \alpha_{v_2}^{\mathcal{G}})$ can be formulated as
\begin{equation*}
    (\alpha_{v_1}^{\mathcal{G}}, \alpha_{v_2}^{\mathcal{G}}) = \mathrm{attn}_{\textit{v}}(\mathbf{X}_{v_1}^{\mathcal{G}}, \mathbf{X}_{v_2}^{\mathcal{G}}),
\end{equation*}
where $\mathrm{attn}_{\textit{v}}$ denotes the layer that implements the view-level attention, and $\mathbf{X}_{v_i}^{\mathcal{G}}$ is the node embeddings from view $v_i$ in subgraph $\mathcal{G}$.
To learn the importance of each view, we first transform view-specific embedding through a fully connected layer, then we calculate the similarity between transformed embedding and a view-level attention vector $\mathbf{q_{\mathcal{G}}}$. We then take the average importance of all the view-specific node embedding as the importance of each view. The importance of each view, denoted as $w_{v_i}$, can be formulated as:
\begin{equation*}
w_{v_i}=\frac{1}{|\mathcal{V}_{\mathcal{G}}|}\sum_{j \in \mathcal{V}_{\mathcal{G}} } \mathbf{q_{\mathcal{G}}^{\mathrm{T}} } \cdot \tanh(\mathbf{W} \cdot \mathbf{x}_{v_i j}^{\mathcal{G}} + \mathbf{b}),
\end{equation*}
where $\mathbf{q}_{\mathcal{G}}$ is the view-level attention vector for each view of feature, $\mathcal{V}_{\mathcal{G}}$ is the nodes of subgraph $\mathcal{G}$, and $\mathbf{x}_{v_i j}^{\mathcal{G}}$ is the embedding of node $j$ in subgraph $\mathcal{G}$ from view $v_i$. Then the weight of each view in subgraph $\mathcal{G}$ can be calculated by
\begin{equation*}
\alpha_{v_i}=\frac{\exp(w_{v_i})}{\exp(w_{v_1}) + \exp(w_{v_2})}.
\end{equation*}
It reflects the importance of each view in our spoiler detection task. Then the fused embeddings of different views can be shown as:
\begin{equation*}
    \mathbf{X}^{\mathcal{G}} = \alpha_{v_1} \cdot \mathbf{X}_{v_1}^{\mathcal{G}}
                      + \alpha_{v_2} \cdot \mathbf{X}_{v_2}^{\mathcal{G}},
\end{equation*}
Thus we get the subgraph-specific node embedding, denoted as $\mathbf{X}^K, \mathbf{X}^M, \mathbf{X}^U$.

We then conduct subgraph-level attention to facilitate the flow of information between the three information sources. Generally, nodes in different subgraphs only contain information from one subgraph. To learn a more comprehensive representation and facilitate the flow of information between subgraphs, we enable the information exchange across various subgraphs using the movie nodes and the review nodes, both appearing in two subgraphs, as the information exchange ports. Specifically, we propose a novel subgraph-level attention to automatically learn the weight of each subgraph and fuse the information learned for different subgraphs. To be specific, the learned weight of each subgraph $(\boldsymbol{\beta}_K, \boldsymbol{\beta}_M, \boldsymbol{\beta}_U)$ can be computed as: 
\begin{equation*}
   (\boldsymbol{\beta}_K, \boldsymbol{\beta}_M, \boldsymbol{\beta}_U) = \mathrm{attn}_\textit{g}(\mathbf{X}^K,\mathbf{X}^M,\mathbf{X}^U),
\end{equation*}
where $\mathrm{attn}_{\textit{g}}$ denotes the subgraph-level attention layer. To learn the importance of each subgraph, we transform subgraph-specific embedding through a feedforward layer and then calculate the similarity between transformed embedding and a subgraph-level attention vector $\mathbf{q}$. Furthermore, we take the average importance of all the subgraph-specific node embedding as the importance of each subgraph. Taking $\mathcal{G}^K$ and $\mathcal{G}^M$ as an example, the shared nodes of these two subgraphs are movie nodes. The importance of each subgraph, denoted as $w^K,w^M$, can be formulated as:
\begin{equation*}
\begin{aligned}
w^V = \frac{1}{|\mathcal{V}_\textit{mv}^{V}|}\sum_{j \in \mathcal{V}_\textit{mv}^{V}} \mathbf{q}^\mathrm{T} \cdot \tanh(\mathbf{W} \cdot \mathbf{x}_j^{V} + \mathbf{b}) \\
\end{aligned}
\end{equation*}
where $V \in \{K,M\}$,  $\mathbf{q}$ is the subgraph-level attention vector for each subgraph. Then the weight of each subgraph can be shown as:
\begin{equation*}
\resizebox{0.48\textwidth}{!}
{$
\beta^{K}=\frac{\exp(w^K)}{\exp(w^K) + \exp(w^M)},\:
\beta^{M}=\frac{\exp(w^M)}{\exp(w^K) + \exp(w^M)}
$}
\end{equation*}

After obtaining the weight, the subgraph-specific embedding can be fused, formulated as:
\begin{equation*}
    \mathbf{X}_\textit{mv} = \beta^K \cdot \mathbf{X}_\textit{mv}^{K} + \beta^M \cdot \mathbf{X}_{\textit{mv}}^{M} 
\end{equation*}
Similarly, for review nodes, we can get the fused representation $\mathbf{X}_{\textit{rv}}$.
Our proposed subgraph-level attention enables the information to flow across different views and subgraphs.

\subsection{\textbf{Overall Interaction}}
One layer of our proposed \ourmethod{} layer, however, cannot enable the information interaction between all information sources (e.g. the user-review subgraph and the knowledge subgraph). In order to further facilitate the interaction of the information provided by each view in each subgraph, we employ $\ell$ \ourmethod{} layers for node representation learning. The representation of movie nodes and review nodes is updated after each layer, incorporating information provided by different views and neighboring subgraphs. This process can be formulated as follows:
\begin{equation*}
    \mathbf{X}^{(i)} = \ourmethod{}(\mathbf{X}^{(i-1)}), 
\end{equation*}
where
\begin{equation*}
\resizebox{0.48\textwidth}{!}
{$
\mathbf{X}^{(i)}=[\mathbf{X}_k^{\mathcal{G^K}(i)},\mathbf{X}_s^{\mathcal{G^K}(i)},\mathbf{X}_m^{\mathcal{G^M}(i)},\mathbf{X}_s^{\mathcal{G^M}(i)},\mathbf{X}_m^{\mathcal{G^U}(i)},\mathbf{X}_s^{\mathcal{G^U}(i)}]
$}
\end{equation*}
We use $\mathbf{h}^{(i)}$ to denote the representation of reviews after adopting the $\textit{i}$-th \ourmethod{} layer.

\subsection{Learning and Optimization}

After a total of $\ell$ \ourmethod{} layers, we obtain the final movie review node representation denoted as $\mathbf{h}^{(\ell)}$. Given a document label $a\in \{\textsc{spoiler}, \textsc{not spoiler}\}$, the predicted probabilities arer calculated as $p(a|\boldsymbol{d}) \propto \mathrm{exp}\big(\mathrm{MLP}_a(\mathbf{h}^{(\ell)})\big)$. We then optimize \ourmethod{} with the cross entropy loss function. At inference time, the predicted label is $\mathrm{argmax}_a p(a|\boldsymbol{d})$.

\section{Experiment}

\subsection{Experiment Settings}
\noindent\textbf{Datasets.} 
We evaluate \ourmethod{} and baselines on two spoiler detection datasets:

\begin{itemize}[leftmargin=*]
    \item \textbf{LCS} is our proposed large-scale automatic spoiler detection dataset. We randomly create a 7:2:1 split for training, validation, and test sets.
    \item \textbf{Kaggle} is a publicly available movie review dataset presented in a Kaggle challenge \citep{misra2019imdb}. 
    We present more details about this dataset in Appendix \ref{sec:dataset_detail}.
\end{itemize} 

\begin{table*}[t]
    \centering
        \caption{Accuracy, AUC, and binary F1-score of \ourmethod{} and three types of baseline methods on two spoiler detection datasets. We run all experiments \textbf{five times} to ensure a consistent evaluation and report the average performance as well as standard deviation. \ourmethod{} consistently outperforms  the three types of methods on both benchmarks.  * denotes that the results are significantly better than the second-best under the student t-test.}
    \begin{adjustbox}{max width=1\textwidth}
    \begin{tabular}{lc c c c c c c c c c}
         \toprule[1.5pt]
         \multirow{2}{*}{\textbf{Model}} &  \multicolumn{3}{c}{\textbf{Kaggle}} & \multicolumn{3}{c}{\textbf{LCS}}\\
         \cmidrule(lr){2-4} \cmidrule(lr){5-7}
         & \textbf{F1} & \textbf{AUC} & \textbf{Acc}  & \textbf{F1} & \textbf{AUC} & \textbf{Acc} \\ 
         \midrule[0.75pt]
         
         \textsc{BERT} \citep{devlin2018bert} & $44.02$ ~\small($\pm1.09$) & $63.46$ ~\small($\pm0.46$) & $77.78$ ~\small($\pm0.09$) & $46.14$ ~\small($\pm2.84$) & $64.82$ ~\small($\pm1.36$) & $79.96$ ~\small($\pm0.38$)  \\
          \textsc{RoBERTa} \citep{liu2019roberta} & $50.93$ ~\small($\pm0.76$) & $66.94$ ~\small($\pm0.40$) & $79.12$ ~\small($\pm0.10$) & $47.72$ ~\small($\pm0.44$) & $65.55$ ~\small($\pm0.22$) & $80.16$ ~\small($\pm0.03$) \\
          \textsc{BART} \citep{lewis2020bart} & $46.89$ ~\small($\pm1.55$) & $64.88$ ~\small($\pm0.71$) & $78.47$ ~\small($\pm0.06$) & $48.18$ ~\small($\pm1.22$) & $65.79$ ~\small($\pm0.62$) & $80.14$ ~\small($\pm0.07$) \\
          \textsc{DeBERETa} \citep{he2021debertav3} & $49.94$ ~\small($\pm1.13$) & $66.42$ ~\small($\pm0.59$) & $79.08$
          ~\small($\pm0.09$) & $47.38$ ~\small($\pm2.22$) & $65.42$ ~\small($\pm1.08$) & $80.13$ ~\small($\pm0.08$) \\
          \midrule[0.75pt]
          \textsc{GCN} \citep{kipf2017semisupervised} & $59.22$ ~\small($\pm1.18$) & $71.61$ ~\small($\pm0.74$) &  $82.08$ ~\small($\pm0.26$) & $62.12$ ~\small($\pm1.18$) & $73.72$ ~\small($\pm0.89$) & $83.92$ ~\small($\pm0.23$) \\
          \textsc{R-GCN} \citep{schlichtkrull2018modeling} & $\underline{63.07}$ ~\small($\pm0.81$) & $\underline{74.09}$ ~\small($\pm0.60$) & $\underline{82.96}$ ~\small($\pm0.09$) & $\underline{66.00}$ ~\small($\pm0.99$) & $\underline{76.18} $ ~\small($\pm0.72$) & $\underline{85.19} $ ~\small($\pm0.21$) \\
            \textsc{SimpleHGN} \citep{lv2021we} & $60.12$ ~\small($\pm1.04$) & $71.61$ ~\small($\pm0.74$) & $82.08$ ~\small($\pm0.26$) & $63.79$ ~\small($\pm0.88$) & $74.64$ ~\small($\pm0.64$) & $84.66$ ~\small($\pm1.61$) \\
           \midrule[0.75pt]
           \textsc{DNSD} \citep{chang2018deep} & $46.33$ ~\small($\pm2.37$) & $64.50$ ~\small($\pm1.11$) & $78.44$ ~\small($\pm0.12$) & $44.69$ ~\small($\pm1.63$) & $64.10$ ~\small($\pm0.74$) & $79.76$ ~\small($\pm0.08$)  \\
           \textsc{SpoilerNet} \citep{wan-etal-2019-fine}  & $57.19$ ~\small($\pm0.66$) & $70.64$ ~\small($\pm0.44$) & $79.85$ ~\small($\pm0.12$) & $62.86$ ~\small($\pm0.38$) & $74.62$ ~\small($\pm0.09$) & $83.23$ ~\small($\pm1.63$) \\        

          \midrule[0.75pt]
          \ourmethod{} (Ours) & $\textbf{65.08}^*$ ~\small($\pm0.69$) & $\textbf{75.42}^*$ ~\small($\pm0.56$) & $\textbf{83.59}^*$ ~\small($\pm0.11$) & $\textbf{69.22}^*$ ~\small($\pm0.61$) & $\textbf{78.26}^*$ ~\small($\pm0.63$) & $\textbf{86.37}^*$ ~\small($\pm0.08$) \\
       
         \bottomrule[1.5pt]
    \end{tabular}
    \end{adjustbox}
    \label{tab:1}
\end{table*}

\noindent\textbf{Baselines.}
We compare \ourmethod{} against 9 baseline methods in three categories: pretrained language models, GNN-based models, and task-specific baselines. For pretrained language models, we evaluate BERT \citep{devlin2018bert}, RoBERTa \citep{liu2019roberta}, BART \citep{lewis2020bart}, and DeBERETa \citep{he2021debertav3}. For GNN-based models, we evaluate GCN \citep{kipf2017semisupervised}, R-GCN \citep{schlichtkrull2018modeling}, and SimpleHGN \citep{lv2021we}. For task-specific baselines, we evaluate DNSD \citep{chang2018deep} and SpoilerNet \citep{wan-etal-2019-fine}.

\begin{figure}[t]
	\centering
	\includegraphics[scale=0.42]{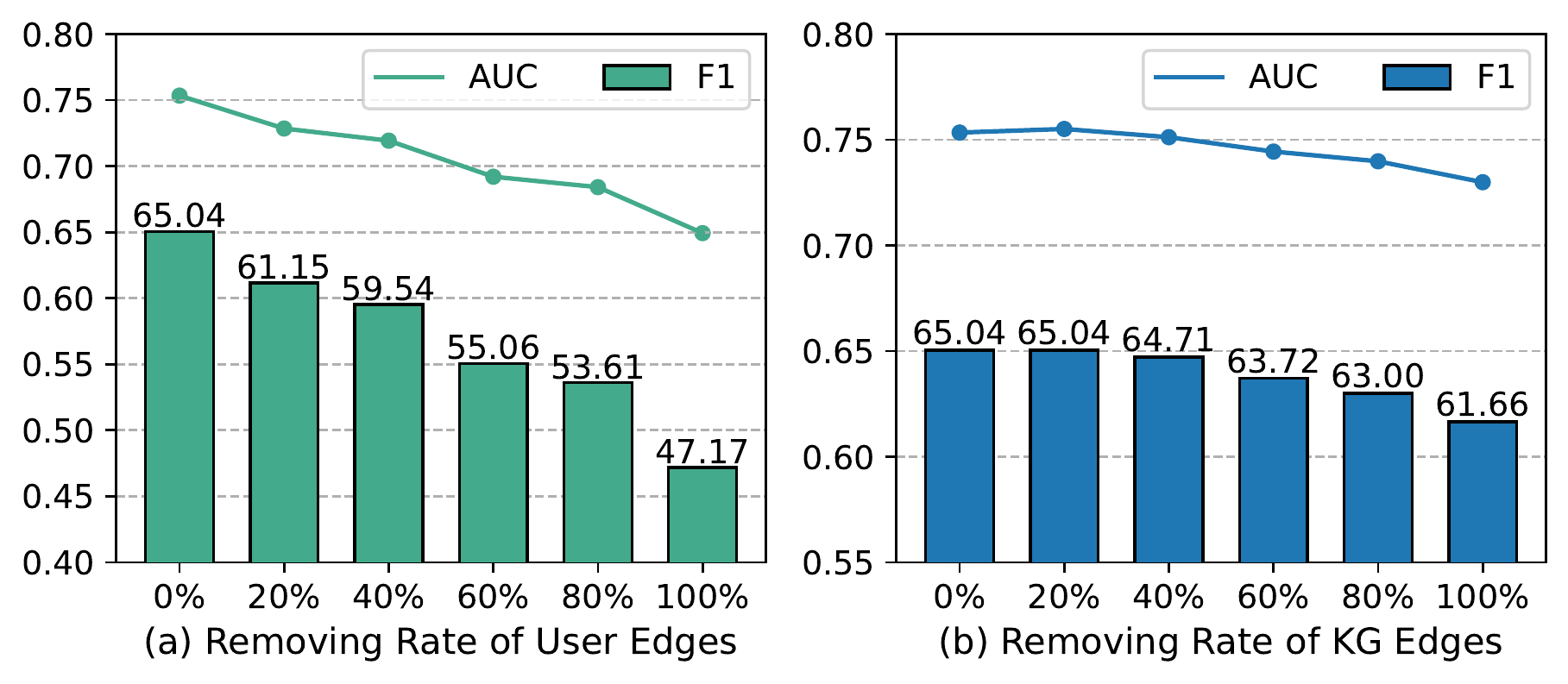}

     	\caption{
    \ourmethod{} performance when randomly removing the edges in the user interaction network and external knowledge subgraph. Performance declines with the gradual edge ablations, indicating the contribution of external knowledge and user networks.
	}
	\label{del}
\end{figure}

\subsection{Overall Performance}
Table \ref{tab:1} presents the performance of \ourmethod{} baseline methods on the two datasets. \textbf{Bold} and \underline{underline} indicate the best and second best performance. Table \ref{tab:1} demonstrates that:
\begin{itemize}[leftmargin=*]
    \item \ourmethod{} achieves state-of-the-art on both datasets, outperforming all baselines by at least 2.01 in F1-score. This demonstrates that our various technical contributions, such as incorporating external knowledge and user networks, multi-view feature extraction, and the cross-context information exchange mechanism, resulted in a more accurate and robust spoiler detection system.
    
    \item Graph-based models are generally more effective than other types of baselines. This suggests that in addition to the textual content of reviews, graph-based modeling could bring in additional information sources, such as external knowledge and user interactions, to enable better grounding for spoiler detection.

    \item Among the two task-specific baselines, SpoilerNet \citep{wan-etal-2019-fine} outperforms DNSD \citep{chang2018deep}, in part attributable to the introduction of the user bias. Our method further incorporates external knowledge and user networks while achieving better performance, suggesting that robust spoiler detection requires models and systems to go beyond the mere textual content of movie reviews.

\end{itemize}


\subsection{External Knowledge and User Networks}
We hypothesize that external movie knowledge and user interactions on movie review websites are essential in spoiler detection, providing more context and grounding in addition to the textual content of movie reviews. To further examine their contributions in \ourmethod{}, we randomly remove 20\%, 40\%, 60\%, 80\%, or 100\% edges of the knowledge subgraph and user-review subgraph, creating settings with reduced knowledge and user information. We evaluate \ourmethod{} with these ablated graphs on the Kaggle dataset and present the results in Figure \ref{del} (a). It is illustrated that the performance drops significantly (about 10\% in F1-score when removing 60\% of the edges) when we increase the number of removed edges in the user-review subgraph, suggesting that the user interaction network plays an important role in the spoiler detection task. As for the knowledge subgraph, the F1-score drops by 3.38\% if we remove the whole knowledge subgraph, indicating that external knowledge is helpful in identifying spoilers. Moreover, it can be observed in Figure\ref{del} (b) that the F1-score and AUC only dropouts slightly when removing part of the edges in the knowledge subgraph. This illustrates the robustness of \ourmethod{}, as it can achieve relatively high performance while utilizing a subset of movie knowledge.

\begin{table}[t]
    \centering
    \caption{Ablation study concerning multi-view data and the graph structure on Kaggle Dataset. The semantic view, knowledge view, and meta view are denoted as S, K, and M respectively. The knowledge subgraph, movie-review subgraph, and user-review subgraph are denoted as $\mathcal{G}^K$, $\mathcal{G}^M$ and $\mathcal{G}^U$.}
    \label{tab:ablation}
    \begin{adjustbox}{max width=0.48\textwidth}
    \begin{tabular}{cc c c c c}
         \toprule[1.5pt]
          {\textbf{Category}} & {\textbf{Setting}} & {\textbf{F1}} & {\textbf{AUC}} & {\textbf{Acc}} \\
         \midrule[0.75pt]
            \multirow{5}{*}{\shortstack{\textbf{multi-view}}} & -w/o S & 38.47 & 61.37 & 78.15 \\
            & -w/o K & 62.13 & 73.46 & 82.73 \\
            & -w/o M & 52.99 & 68.07 & 79.46 \\
            & -w/o O, K & 40.05 & 61.97 & 78.25 \\
            & -w/o O, M & 56.44 & 70.05 & 80.66 \\
            
            \midrule[0.75pt]
            
           \multirow{4}{*}{\shortstack{\textbf{graph}\\ \textbf{structure} }} & -w/o $\mathcal{G}^K$ & 61.66 & 72.99 & 83.12 \\
            & -w/o $\mathcal{G}^U$ & 47.17 & 64.93 & 78.00 \\
            & -w/o $\mathcal{G}^M$, $\mathcal{G}^K$ & 56.54 & 69.98 & 81.71 \\
            & -w/o $\mathcal{G}^M$, $\mathcal{G}^K$ & 46.65 & 64.89 & 78.03 \\
             \midrule[0.75pt]
             \textbf{ours} & \ourmethod{} & \textbf{65.08} & \textbf{75.42} & \textbf{83.59} \\
         \bottomrule[1.5pt]
    \end{tabular}
    \end{adjustbox}
\end{table}
 
\subsection{Ablation Study}

In order to study the effect of different views of data, we remove them individually and evaluate variants of our proposed model on the Kaggle Dataset. We further remove some parts of the graph structure to investigate, Finally, we replace our attention mechanism with simple fusion methods to evaluate the effectiveness of our fusion method.

\paragraph{\textbf{Multi-View Study}}

\begin{table}[t]
    \centering
        \caption{Model performance on Kaggle when our attention mechanism is replaced with simple fusion methods.}
    \label{tab:3}
    \begin{adjustbox}{max width=0.48\textwidth}
    \begin{tabular}{cc c c c c c c}
         \toprule[1.5pt]

         {\textbf{View-level}} & {\textbf{Subgraph-level}} & {\textbf{F1}} & {\textbf{AUC}}&
         {\textbf{Acc}}
         \\
         \midrule[0.75pt]
            Ours  & Max-pooling & 53.73 & 68.50 & 79.29 \\ 
            Ours  & Mean-pooling & 62.27 & 73.40 & 83.23 \\
            Ours  & Concat & 61.07 & 72.63 & 82.97 \\
            Max-pooling  & Ours & 63.19 & 74.21 & 82.86 \\
            Mean-pooling  & Ours & 63.60 & 74.36 & 83.30 \\
            Concat  & Ours & 62.90 & 74.00 & 82.83 \\
            \midrule[0.75pt]
            Ours & Ours & \textbf{65.08} & \textbf{75.42} & \textbf{83.59} \\
         \bottomrule[1.5pt]
    \end{tabular}
    \end{adjustbox}
\end{table}

We report the binary F1-Score, AUC, and Acc of the ablation study in Table \ref{tab:ablation}. Among the multi-view data, semantic view data is of great significance as AUC and F1-score drop dramatically when it is discarded. We can see that discarding the external knowledge view or removing the knowledge subgraph reduces the F1-score by about 3\%, indicating that the external knowledge of movies is helpful to the spoiler detection task. However,  external knowledge doesn't show the same importance as the directly related semantic view or meta view. We believe this is because the external knowledge is not directly related to review documents, so it can only provide auxiliary help to the spoiler detection task. 

\paragraph{\textbf{Graph Structure Study}}
As illustrated in Table \ref{tab:ablation}, after removing the user-review subgraph, the reduced model performs poorly, with a drop of 18\% in F1. This demonstrates that the user interaction network is necessary for spoiler detection. 

\paragraph{Aggregation and Interaction Study}
In order to study the effectiveness of the hierarchical mechanism that enables the interaction between views and sub-graphs, we replace the two components of our hierarchical attention with other operations and evaluate them on the Kaggle Dataset. Specifically, we compare our attention module with concatenation, max-pooling, and average-pooling.

In Table \ref{tab:3} we report the binary F1-score, AUC, and Acc. We can see that our approach beats the eight variants in all metrics. It is evident that our approach can aggregate and fuse multi-view data more efficiently than simple fusion methods.

\subsection{Qualitative Analysis}
We conduct qualitative analysis to investigate the role of external movie knowledge and social networks for spoiler detection. As shown in Table \ref{tab:qualitative}, with the guide of external knowledge and user networks, \ourmethod{} successfully makes the correct prediction while baseline models fail. Specifically, in the first case, the user is a fan of Kristen Wiig. Guided by the information from the social network, \ourmethod{} finds that the user often posted spoilers related to the film star, and finally predicts that the review is a spoiler. In the second case, the user mentioned something done by the director of the movie. With the help of movie knowledge, it can be easily distinguished that what the director has done reveals nothing of the plot. 

\begin{table*}[t]
    \centering
    \caption{Examples of the performance of three baselines and \ourmethod{}. Underlined parts indicate the plots.}
    \label{tab:qualitative}
    \begin{adjustbox}{max width=1\textwidth}
    \begin{tabular}{p{5in}llllll}
         \toprule[1.5pt]
         {\textbf{Review Text}} & {\textbf{Label}} & {\textbf{DeBERTa}} &
         {\textbf{R-GCN}} & {\textbf{SpoilerNet}} & {\textbf{\ourmethod{}}}
         \\
         \midrule[0.75pt]
         Kristen Wiig is the only reason I wanted to see this movie, and she is insanely hilarious! (...)
         Wiig plays Annie, (...) \underline{becomes jealous of Lillian's new rich friend,} \underline{Helen. Annie slowly goes crazy and constantly competes against Helen (...)}
        & \multirow{2}{*}{True} & \multirow{3}{*}{\shortstack{False \\ \\  \XSolidBrush}} & \multirow{3}{*}{\shortstack{False \\ \\ \XSolidBrush}} & \multirow{3}{*}{\shortstack{False \\ \\ \XSolidBrush}} & \multirow{3}{*}{\shortstack{True \\ \\ \CheckmarkBold}} \\
        \midrule[0.75pt]
        The new director was horrible. Not even comparable to Chris Columbus. He changed the entire format of the school (...) why was there a deer next to harry across the lake, he didn't mention that and yet he still put the deer in the movie (...)
         & \multirow{2}{*}{\shortstack{False}} & \multirow{3}{*}{\shortstack{True \\ \\  \XSolidBrush}} & \multirow{3}{*}{\shortstack{True \\ \\ \XSolidBrush}} & \multirow{3}{*}{\shortstack{True \\ \\ \XSolidBrush}} & \multirow{3}{*}{\shortstack{False \\ \\ \CheckmarkBold}} \\
        \midrule[0.75pt]
        (...) This scene involves Harry getting bombarded by ugly, little squid like creatures and is awe inspiring. And more happens. Harry is having a certain dream over and over again. Lord Voldemort wants to return and he does.
        & \multirow{2}{*}{\shortstack{True}} & \multirow{3}{*}{\shortstack{False \\ \\ \XSolidBrush}} & \multirow{3}{*}{\shortstack{False \\ \\ \XSolidBrush}} & \multirow{3}{*}{\shortstack{False \\ \\ \XSolidBrush}} & \multirow{3}{*}{\shortstack{True \\ \\ \CheckmarkBold}} \\
        \midrule[0.75pt]

        (...) I remember that for four years in high school, I was a high school nerd/loner, and I liked it. I was shy, I was socially awkward, and I was one of those guys who happened to have a thing for one of the popular girls (...) & \multirow{2}{*}{\shortstack{False}} & \multirow{3}{*}{\shortstack{True \\ \\  \XSolidBrush}} & \multirow{3}{*}{\shortstack{True \\ \\ \XSolidBrush}} & \multirow{3}{*}{\shortstack{True \\ \\ \XSolidBrush}} & \multirow{3}{*}{\shortstack{False \\ \\ \CheckmarkBold}} \\
         \bottomrule[1.5pt]
    \end{tabular}
    \end{adjustbox}
\end{table*}

\section{Related Work}
\paragraph{Automatic Spoiler Detection}
Automatic spoiler detection aims to identify spoiler reviews in domains such as television \citep{10.5555/2655780.2655825}, books \citep{wan-etal-2019-fine}, and movies \citep{misra2019imdb, 10.5555/2655780.2655825}. Existing spoiler detection models could be mainly categorized into two types: keyword matching and machine learning models. Keyword matching methods utilize predefined keywords to detect spoilers, for instance, the name of sports teams or sports events \citep{nakamura2007temporal}, or the name of actors \citep{golbeck2012twitter}. This type of method requires keywords defined by humans, and cannot be generalized to various application scenarios. Early neural spoiler detection models mainly leverage topic models or support vector machines with handcrafted features. \citet{guo-ramakrishnan-2010-finding} use bag-of-words representation and 
LDA-based model to detect spoilers, \citet{jeon2013don} utilize SVM classification with four extracted features, while \citet{10.5555/2655780.2655825} incorporate lexical features and meta-data of the review subjects (e.g., movies and books) in an SVM classifier. Later approaches are increasingly neural methods: \citet{chang2018deep} focus on modeling external genre information based on GRU and CNN, while \citet{wan-etal-2019-fine} introduce item-specificity and bias and utilizes bidirectional recurrent neural networks (bi-RNN) with Gated Recurrent Units (GRU).
A recent work \citep{chang-etal-2021-killing} leverages dependency relations between context words in sentences to capture the semantics using graph neural networks. 

While existing approaches have made considerable progress for automatic spoiler detection, it was previously underexplored whether review text itself is sufficient for robust spoiler detection, or whether more information sources are required for better task grounding. In this work, we make the case for incorporating external film knowledge and user activities on movie review websites in spoiler detection, advancing the field through both resource curation and method innovation, presenting a large-scale dataset LCS, an up-to-date movie knowledge base UKM, and a state-of-the-art spoiler detection approach \ourmethod{}.

\paragraph{Graph-Based Social Text Analysis}
Graphs and heterogeneous information networks are playing an important role in the analysis of texts and documents on news \citep{mehta-etal-2022-tackling} and social media \citep{hofmann-etal-2022-modeling}. In these approaches, graphs and graph neural networks are adopted to represent and encode information in addition to textual content, such as social networks \citep{nguyen2020fang}, external knowledge graphs \citep{zhang-etal-2022-kcd}, social context \citep{mehta-etal-2022-tackling}, and dependency relations between context words \citep{chang-etal-2021-killing}. With the help of additional information sources, these graph-based approaches enhance representation quality by capturing the rich social interactions \citep{nguyen2020fang},  infusing knowledge reasoning into language representations \citep{zhang-etal-2022-kcd}, and reinforcing nodes’
representations interactively \citep{mehta-etal-2022-tackling}. As a result, graph-based social text analysis approaches have advanced the state-of-the-art on various tasks such as misinformation detection \citep{zhang-etal-2022-kcd}, stance detection \citep{liang-etal-2022-jointcl}, propaganda detection \citep{vijayaraghavan-vosoughi-2022-tweetspin}, sentiment analysis \citep{chen-etal-2022-enhanced}, and fact verification \citep{arana-catania-etal-2022-natural}. Motivated by the success of existing graph-based models, we propose \ourmethod{} to incorporate external knowledge bases and user networks on movie review platforms through graphs and graph neural networks.

\section{Conclusion}
We make the case for incorporating external knowledge and user networks on movie review websites for robust and well-grounded spoiler detection. Specifically, we curate LCS, the largest spoiler detection dataset to date; we construct UKM, an up-to-date knowledge base of the film industry; we propose \ourmethod{}, a state-of-the-art spoiler detection system that takes external knowledge and user interactions into account. Extensive experiments demonstrate that \ourmethod{} achieves state-of-the-art performance on two datasets while showcasing the benefits of incorporating movie knowledge and user behavior in spoiler detection. We leave it for future work to further check the labels in the LCS dataset.

\clearpage

\section*{Limitations}
\label{sec:limit}
We identify two key limitations:

\begin{itemize}[leftmargin=*]
    \item \textbf{\ourmethod{}} utilizes widely-adopted RGCN to model each subgraph, while there are more up-to-date heterogeneous graph algorithms like HGT \citep{hgt}, SimpleHGN \citep{lv2021we}. We plan to conduct experiments that replace RGCN with other heterogeneous graph algorithms. Besides, considering the subgraph structure of \ourmethod{}, we will test different heterogeneous graph algorithm settings in each subgraph to find out the most efficient algorithm for each subgraph.
    \item \textbf{LCS} is constructed based on IMDB, and the spoiler annotation is based on user self-report. Hence, it is likely that some label is false. In the next step of our work, we will check the labels with the help of experts and weak supervised learning strategy \citep{zhou2018brief}.
\end{itemize}

\section*{Ethics Statement}
We envision \ourmethod{} as a pre-screening tool and not as an ultimate decision-maker. Though achieving the state-of-the-art, \ourmethod{} is still imperfect and needs to be used with care, in collaboration with human moderators to monitor or suspend suspicious
movie reviews. Moreover, \ourmethod{} may inherit the biases of its constituents, since it is a combination of datasets and models. For instance, pretrained language models could encode undesirable social biases and stereotypes \citep{li-etal-2022-herb, nadeem-etal-2021-stereoset}. We leave to future work on how to incorporate the bias detection and mitigation techniques developed in ML research in spoiler detection systems.
Given the nature of the task, the dataset contains potentially offensive language which should be taken into consideration. 

\section*{Acknowledgements}
We would like to thank Bocheng Zou for helping collect the data. We would also
like to thank all LUD Lab members for our collaborative research environment.
Minnan Luo and Qinghua Zheng are supported by the National Key Research and Development Program of China (No. 2022YFB3102600), National Nature Science Foundation of China (No. 62192781, No. 62272374, No. 62202367, No. 62250009, No. 62137002), Innovative Research Group of the National Natural Science Foundation of China (61721002), Innovation Research Team of Ministry of Education (IRT\_17R86), Project of China Knowledge Center for Engineering Science and Technology,  Project of Chinese academy of engineering ``The Online and Offline Mixed Educational Service System for `The Belt and Road' Training in MOOC China'', and the K. C. Wong Education Foundation.

\bibliographystyle{acl_natbib}
\bibliography{anthology,custom}


\clearpage

\appendix

\section{Dataset Details}
\label{sec:dataset_detail}
We adopt two graph-based spoiler detection datasets, namely Kaggle \citep{misra2019imdb} and our curated LCS. The two datasets are both in English. The publicly available Kaggle dataset only provides incomplete information. Hence, we retrieved cast information based on the movie ids and collected user metadata based on user ids. The statics details of Kaggle after retrieving are listed in Table \ref{tab:kaggle}, and the statics details of our LCS are listed in Table \ref{tab:dataset}.

\begin{table}[t]
    \centering
    \caption{Statistics of the Kaggle Dataset.}
    \begin{adjustbox}{max width=0.48\textwidth}
    \begin{tabular}{c|c|c}
        \toprule[1.5pt]  \textbf{Type}&\textbf{Number}&\textbf{description}\\
        \midrule[1pt]
         review & 573,913 & The posting time is from 1998 to 2018. \\ 
         user & 263,407 & Users that posted these reviews. \\
         movie & 1,572 & The released year is from 1921 to 2018. \\ 
         cast & 7,865 & The cast related to the movies. \\ 
         spoiler & 150,924 & 25.87\% of the reviews are spoilers. \\
         \bottomrule[1.5pt]
    \end{tabular}
    \end{adjustbox}
    \label{tab:kaggle}
\end{table}

\begin{table}[t]
    \centering
    \caption{Details of metadata contained in the dataset.}
    \begin{adjustbox}{max width=0.48\textwidth}
    \begin{tabular}{c|c}
        \toprule[1.5pt]  \textbf{Entity Name}&\textbf{Metadata}\\
        \midrule[1pt]
         Review & time, helpful vote count, total vote count, score \\ 
         User & create at, badge count, review count \\
         Movie & year, isAdult, runtime, rating, vote count \\ 
         Cast & birth year, death year, involved movie count \\
         \bottomrule[1.5pt]
    \end{tabular}
    \end{adjustbox}
    \label{tab:metadata}
\end{table}

\begin{figure*}[ht]
	\centering
	\includegraphics[scale=0.38]{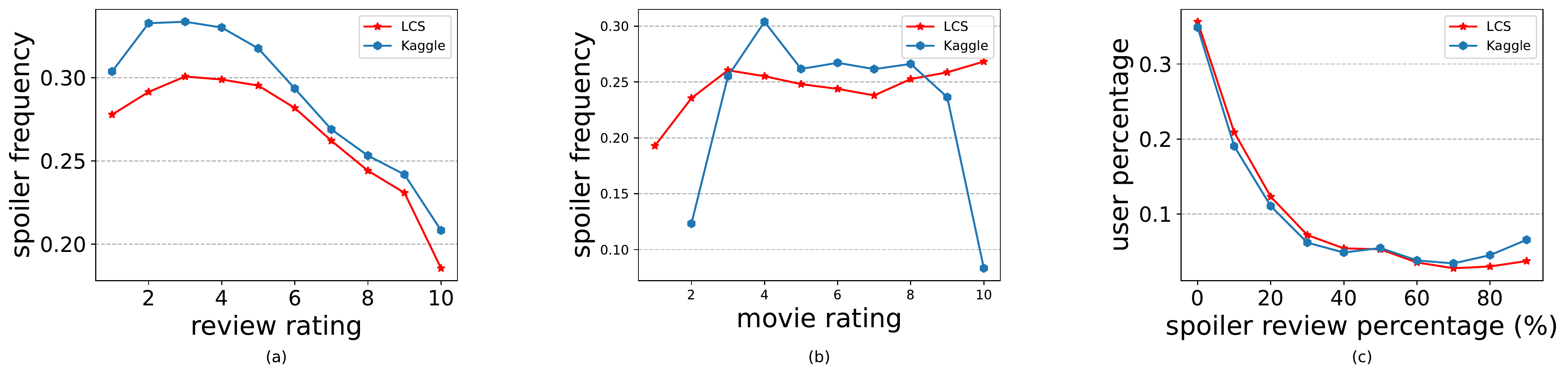}
	\caption{(a) The spoiler frequency of reviews with different ratings; (b) The spoiler frequency of reviews related to movies of different ratings; (c) The percentage of spoilers per user, spoiler review percentage intervals are divided every 10 percent.}
	\label{meta_ana}
\end{figure*}

\subsection{Data Analysis}
\label{sec:data_analysis}
We compare \textbf{LCS} with another popular spoiler detection dataset Kaggle \citep{misra2019imdb} and presents our findings in Figure \ref{meta_ana}. We investigate the correlation between spoilers and individual review scores, overall movie ratings, and the behavior of different users.
Firstly, we investigate the correlation between spoilers and review scores. Figure \ref{meta_ana}(a) shows that whether a review containing spoilers has a strong connection with how well the user considers the movie. Additionally, we find that whether a review contains spoilers is also related to the public opinion of the movie, which is illustrated in Figure \ref{meta_ana}(b). These findings suggest the necessity of leveraging metadata and external knowledge of movies.
In addition, we study the fraction of reviews containing spoilers per user. As illustrated in Figure \ref{meta_ana}(c), the 'spoiler tendency' varies greatly among users. This suggests that it is essential to utilize the user information and how they interact with different movies on review websites.

\subsection{Metadata}
The metadata we collected for both datasets is listed in Table \ref{tab:metadata}.
\label{sec:metadata}

\section{KG Details}
The types of relations, triples, and the number of them are presented in Table \ref{tab:KG_Entity}.
\begin{table}[t]
    \centering
    \caption{Statistics of UKM.}
    \begin{adjustbox}{max width=0.48\textwidth}
    \begin{tabular}{c|c|c}
        \toprule[1.5pt]  
        \textbf{Relation} & \textbf{Triple (head-rel.-tail)}& \textbf{Value} \\
        \midrule[0.75pt]
        show\_in & movie-show\_in-year & 147,191  \\
         rated & movie-rated-rating & 147,191 \\
        genre\_is & movie-genre\_is-genre & 147,191 \\
        is\_director\_of & person-is\_director\_of-movie & 129,483 \\
        is\_actor\_of & person-is\_actor\_of-movie & 379,696 \\
        is\_actress\_of & person-is\_actress\_of-movie & 226,775 \\
        is\_producer\_of & person-is\_producerr\_of-movie & 129,202 \\
        is\_writer\_of & person-is\_writer\_of & 169,024 \\
        is\_editor\_of & person-is\_editor\_of-movie & 49,817 \\
        is\_composer\_of & person-is\_composer\_of-movie & 89,572 \\
        is\_production\_designer\_of & person-is\_production\_designer\_of-movie & 11,838 \\
        is\_archive\_footage\_of & person-is\_archive\_footage\_of-movie & 6,328 \\
        is\_cinematographer\_of & person\-cinematographer\_of-movie & 76,311 \\
        is\_archive\_sound\_of & person-is\_archive\_sound\_of & 205 \\
        is\_self\_of & person-is\_self\_of-movie & 129,483 \\
        
         \bottomrule[1.5pt]
    \end{tabular}
     \end{adjustbox}
    \label{tab:KG_Entity}
\end{table}

\begin{figure}[t]
    \centering

    \label{fig:my_label}
     \begin{subfigure}[b]{0.23\textwidth}
         \centering
         \includegraphics[width=\textwidth]{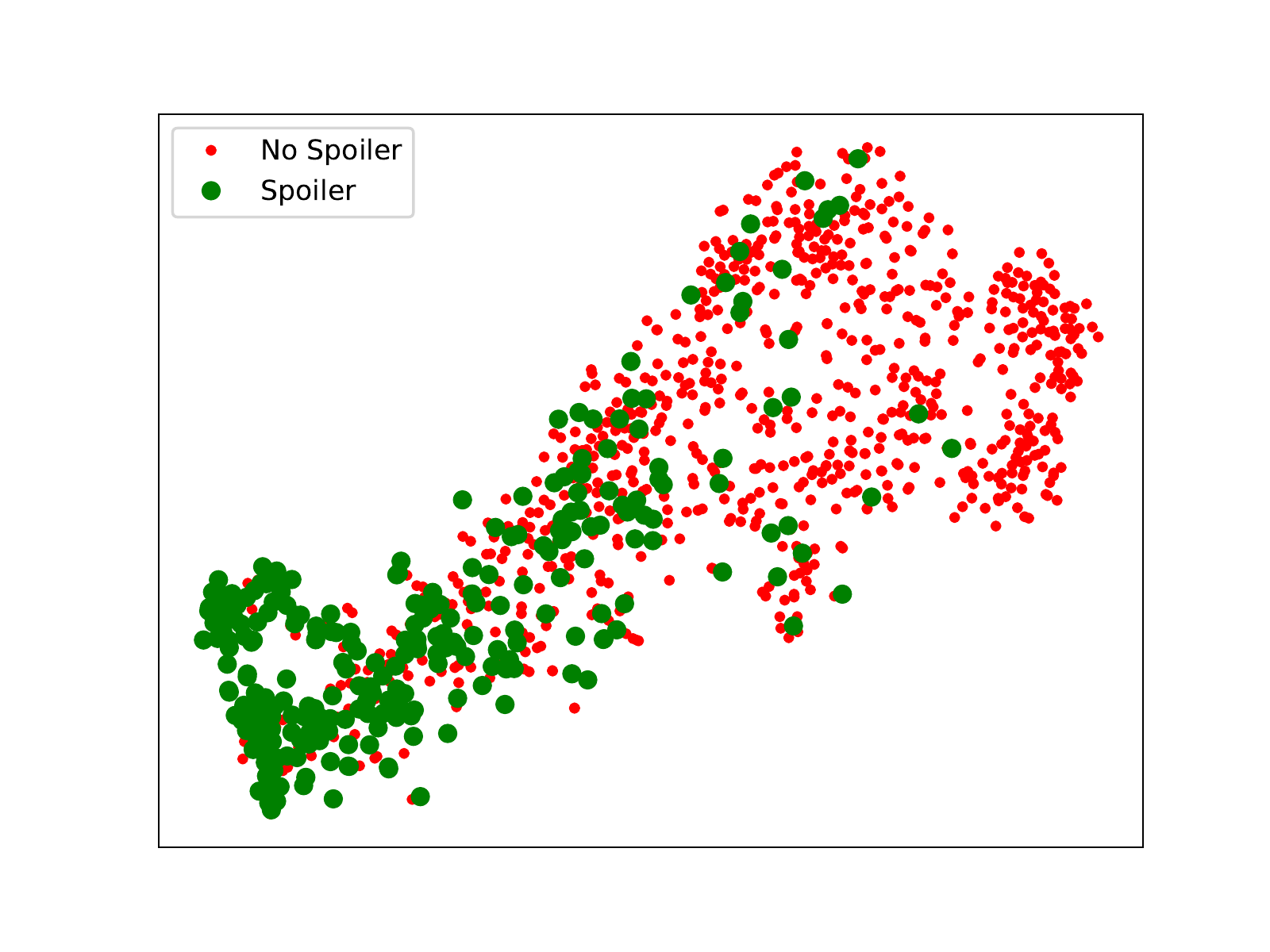}
         \caption{\ourmethod{}.}
         \label{fig:ours}
    \end{subfigure}
    \begin{subfigure}[b]{0.23\textwidth}
         \centering
         \includegraphics[width=\textwidth]{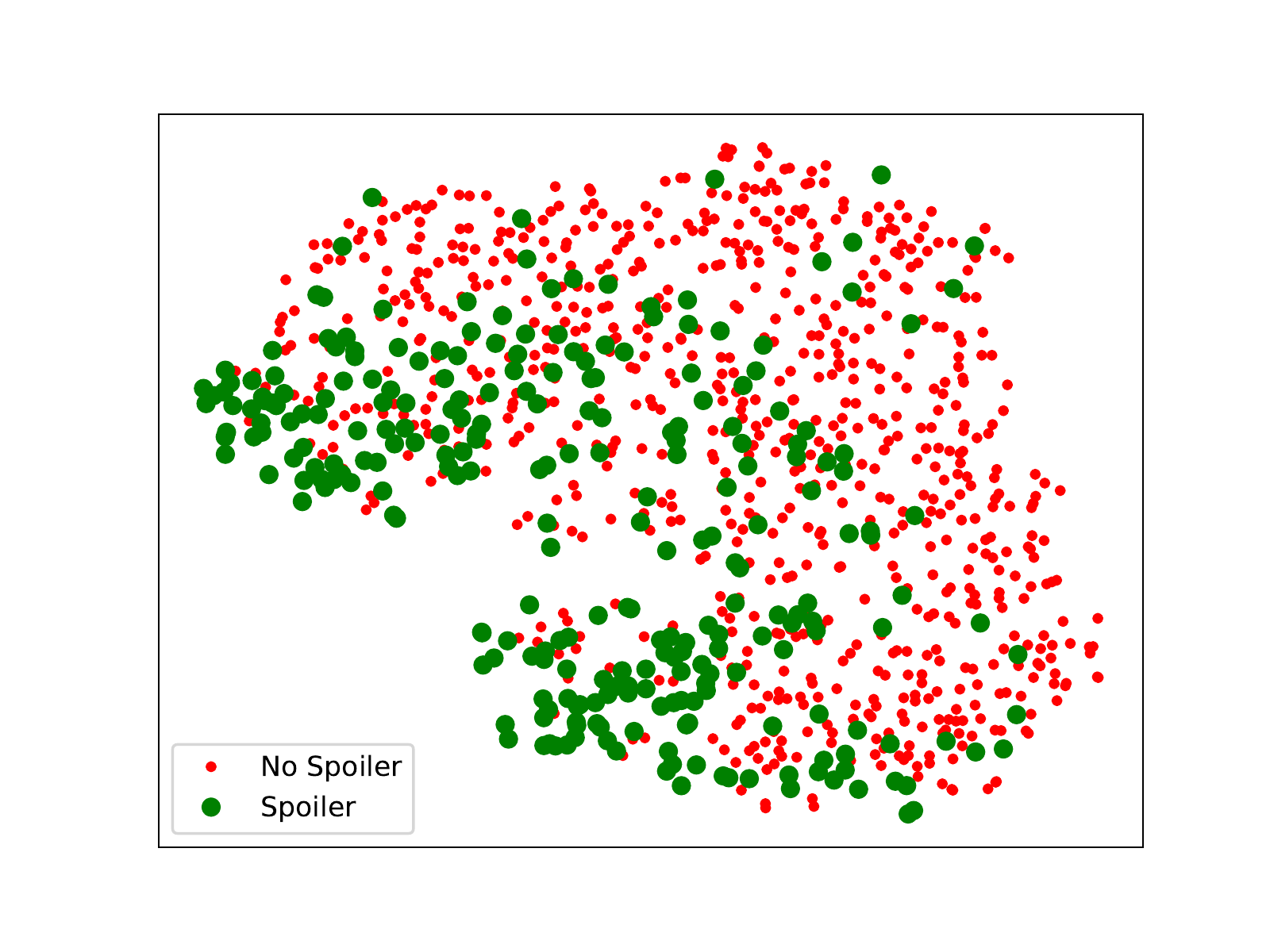}
         \caption{R-GCN.}
         \label{fig:rgcn}
    \end{subfigure}
            \caption{T-SNE visualization of representations of reviews learned by \ourmethod{} and R-GCN.}
\end{figure}

\section{Experiment Details}
\label{sec:exp}
\noindent\textbf{Implementation.} For pre-trained LMs, we utilize the pre-trained model to get the embeddings and transform them through MLPs. For DNSD and SpoilerNet, we follow the settings in their corresponding papers. For GNNs, we combined the three subgraphs into a whole graph and only utilize the semantic view embedding.  We learn a representation for each review, and the representations are passed to an MLP for classification.

\subsection{Baseline Details}
\label{sec:baseline_detail}
We compare \ourmethod{} with pre-trained language models, GNN-based models, and task-specific baselines to ensure a holistic evaluation. For pre-trained language models, we pass the review text to the model, average all tokens, and utilize two fully connected layers to conduct spoiler detection. For GNN-based models, we pass the review text to RoBERTa, averaging all tokens to get the initial node feature. We provide a brief description of each of the baseline methods, in the following.
\begin{itemize}[leftmargin=*]
   \item{\textbf{BERT} \citep{devlin2018bert} is a language model pre-trained on a large volume of natural language corpus with the masked language model and next sentence prediction objectives.}
    \item{\textbf{RoBERTa} \citep{liu2019roberta} improves upon BERT by removing the next sentence prediction task and improves the masking strategies.}
    \item{\textbf{BART} \citep{lewis2020bart} is a transformer encoder-decoder (seq2seq) language model with a bidirectional (BERT-like) encoder and an autoregressive (GPT-like) decoder.}
    \item{\textbf{DeBERTa} \citep{he2021deberta} improves existing language models using disentangled attention and enhanced mask decoder.}

    \item{\textbf{GCN} \citep{kipf2017semisupervised} is short for graph convolutional networks, which enables parameterized message passing between neighbors.}
    \item{\textbf{R-GCN} \citep{schlichtkrull2018modeling} extends GCN to enable the processing of relational networks.}
    \item{\textbf{SimpleHGN} \citep{lv2021we}} is a simple yet effective GNN for heterogeneous graphs inspired by the GAT \citep{velivckovic2018graph}.
    \item{\textbf{DNSD} \citep{chang2018deep} is a
    spoiler detection framework using a CNN-based genre-aware attention mechanism.}
    \item{\textbf{SpoilerNet} \citep{wan-etal-2019-fine} extends the hierarchical attention network (HAN) \citep{yang2016hierarchical} with item-specificity information and item and user bias terms for spoiler detection.}
\end{itemize}

\subsection{Hyperparameter Details}
We present our hyperparameter settings in Table \ref{tab:hyperparameter} to facilitate reproduction. The setting for both datasets is the same. 
\begin{table}[t]
    \centering
        \caption{Hyperparameter settings of \ourmethod{}.}
    \label{tab:hyperparameter}
    \begin{tabular}{l c}
         \toprule[1.5pt] \textbf{Hyperparameter} & \textbf{Value} \\ \midrule[0.75pt]
         GNN input size & 768 \\
         GNN hidden size & 128 \\
         GNN layer (in each \ourmethod{} layer) & 1 \\
         \ourmethod{} layer $L$ & 2\\
         \# epoch & 120 \\
         batch size & 1,024 \\
         dropout & 0.3 \\
         learning rate & 1e-3 \\
         weight decay & 1e-5 \\
         lr\_scheduler\_patience & 5 \\
         lr\_scheduler\_step & 0.1 \\
         Optimizer & AdamW \\
         \bottomrule[1.5pt]
    \end{tabular}

\end{table}

\subsection{Computational Resources}
Our proposed approach has a total of 0.9M learnable parameters. It takes about 10 GPU hours to train our approach on the Kaggle dataset. We train our model on a Tesla V100 GPU. We conduct all experiments on a cluster with 4 Tesla V100 GPUs with 32 GB memory, 16 CPU cores, and 377GB CPU memory.
\subsection{Experiment Runs}
For both datasets that have relatively large scales, we adopt the subsampling skill proposed in \citep{hamilton2017inductive}, which has been successfully used on large graphs \citep{velickovic2019deep}. We conduct our approach and baselines five times on both datasets and report the average F1-score, AUC, and accuracy with standard deviation in Table \ref{tab:1}. For the experiments in Table \ref{tab:ablation}, Table \ref{tab:3}, and Figure \ref{del}, we only report the single-run result in the Kaggle dataset due to the lack of computational resources.

\subsection{Visualization}
To intuitively demonstrate the effectiveness of our representation method, we utilize T-SNE \citep{van2008visualizing} to visualize the representations of movie reviews learned by different models. Specifically, we choose our proposed \ourmethod{} and R-GCN (with the second highest performance) and evaluate them on the validation set of the small dataset. It can be observed in Figure \ref{fig:rgcn} that the learned representations of different kinds are relatively mixed together. In contrast, representations learned by \ourmethod{} show moderate collocation for both groups of reviews. This illustrates that \ourmethod{} yields improved and more comprehensive representation with the effective use of multi-view data and user interaction networks.

\begin{figure*}[ht]
	\centering
	\includegraphics[scale=0.40]{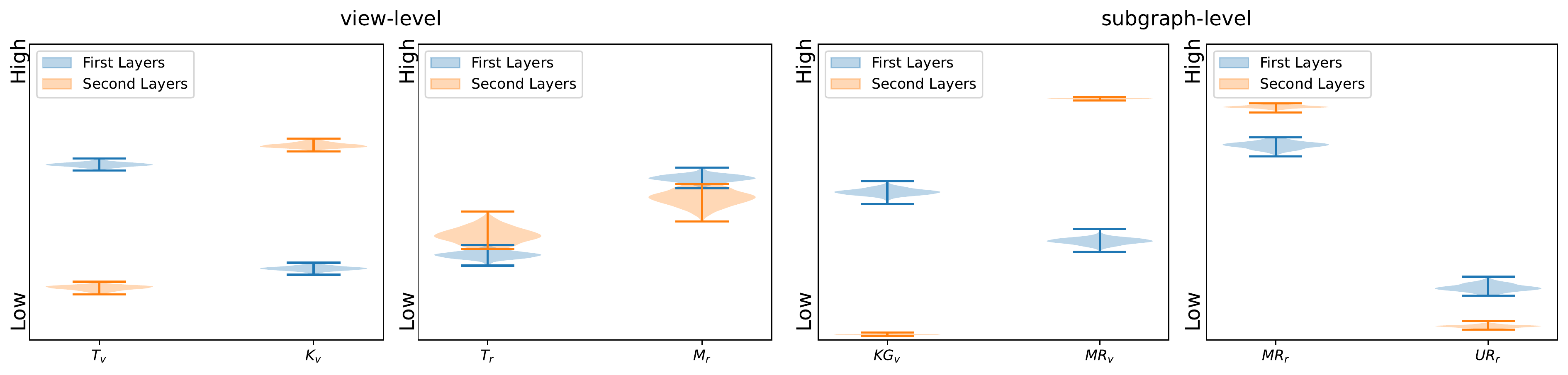}
	\caption{
    Attention weights learned by our hierarchical attention. Subscript $v$, $r$ indicate the public nodes movie and review separately. $T$, $M$, and $K$ refer to the textual view, the meta view, and  the external knowledge view, respectively. This violin plot illustrates the different contributions of each view and subgraph and the process of interaction.
	}
	\label{attn_weight}
\end{figure*}

\subsection{Contribution of Views and Subgraphs}
We introduce semantic, meta, and external knowledge views and utilize user-review, movie-review, and knowledge subgraph structures to represent multi-information. To further study the contribution of different views and sub-graphs. We extract the attention weight from the View-level attention layers and Subgraph-level attention layers and illustrate them in violin plots. We select representative features and present them in Figure \ref{attn_weight}. The four violin plots demonstrate that our proposed hierarchical attention can select the more important features from the variation of attention weight between the first and the second layer, indicating that the contributions of certain representations are varied as they capture features via the graph structure and attention mechanism.


\section{Significance Testing}
To further evaluate \ourmethod{}'s performance on both datasets, we apply one way repeated measures ANOVA test for the results in Table \ref{tab:1}. The result demonstrates that the performance gain of our proposed model is significant on both datasets against the second-best R-GCN on all three metrics with a confidence level of 0.05.

\section{Scientific Artifact Usage}
The \ourmethod{} model is implemented with the help of many widely-adopted scientific artifacts, including PyTorch \citep{paszke2019pytorch}, NumPy \citep{harris2020arraynumpy}, transformers \citep{wolf2020transformers}, sklearn \citep{scikit-learn}, OpenKE \citep{han-etal-2018-openke}, PyTorch Geometric \citep{fey2019pytorchgeometric}. We utilize data from IMDB and following the requirement of IMDB, we acknowledge the source of the data by including the following statement: Information courtesy of
IMDb (https://www.imdb.com). Used with permission. Our use of IMDb data is non-commercial, which is allowed by IMDB. We will make our code and data publicly available to facilitate reproduction and further research.

\end{document}